\documentclass[11pt,a4paper]{article}
\usepackage[hyperref]{acl2017onecolumn}
\usepackage{times}
\usepackage{latexsym}
\usepackage{amssymb}
\usepackage{url}

\usepackage[paper=a4paper, bottom=1.5in,right=0in,left=0in]{geometry}

\usepackage{mathtools}
\usepackage{tabu}
\usepackage{linguex}
\usepackage{pgfplots}
\usepackage{tikz}
\usepackage{booktabs}
\newcommand{\tabitem}{~~\llap{}~~}

\aclfinalcopy 

\begin{document}

\begin{titlepage}
	\begin{center}
		\includegraphics[width=0.4\textwidth]{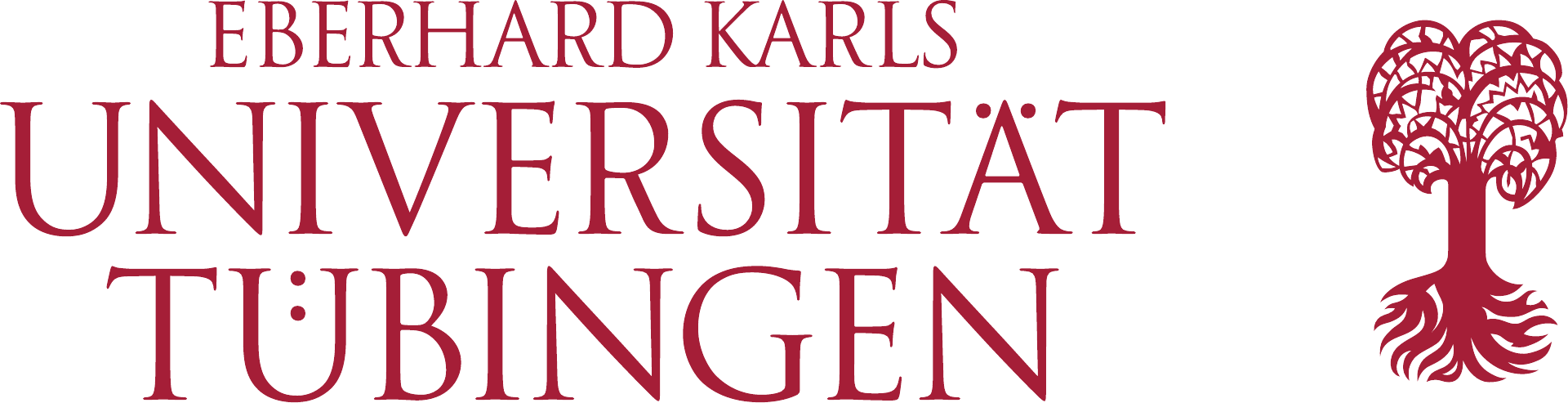}\\
		\vspace*{3cm}
		\Large
		\textbf{Bachelor Thesis}\\
		\vspace*{1.7cm}
		\LARGE
		\textbf{Semi-supervised emotion lexicon expansion\\ with label propagation and specialized word embeddings}
		\vspace{1.7cm}

		\Large
		A thesis presented for the degree of\\
		\textbf{Bachelor of Arts} in \\\textbf{International Studies in Computational Linguistics}

		\vspace{1.7cm}

		\Large \textbf{Mario Giulianelli}\\
		\vspace{0.5cm}
		July 2017

		\vfill

		\Large
		Supervised by Dr Dani\"el de Kok \\
		\vspace{0.5cm}
		Seminar f\"ur Sprachwissenschaft\\
		Universit\"at T\"ubingen\\
		Germany

	\end{center}
\end{titlepage}


	\newgeometry{top=1.2in,bottom=1.2in,right=1.45in,left=1.5in,footskip=.6in}  
	
	\clearpage
	\vspace*{\fill}
	\begin{center}
		\begin{minipage}{0.55\textwidth}
			\renewcommand{\baselinestretch}{1.6}
			\huge
			\textbf{The main contribution of this Bachelor thesis will soon be reported in a standalone paper.}
		\end{minipage}
	\end{center}
	\vfill 
	\clearpage

	\pagenumbering{arabic}
	\pagestyle{plain}
	

\section{Introduction} \label{sec:intro}
As the amount of online data increases, new methods are continuously developed to make sense of the available information. The Web and social networks have allowed anyone with an Internet connection to contribute to this enormous, freely available database. The data is mostly unorganized and text is probably the most abundant unstructured resource. On online platforms, users express their opinions and share their experiences, thus they generate a complex network of mutual influence. Reviews of goods and services, political views and commentaries, as well as recommendations of job applicants exemplify how Web content can impact the decision-making process of consumers, voters, companies, and other organizations \citep{pang2008}.
Therefore, it is not surprising that sentiment analysis and opinion mining are active areas of academic and industrial research nowadays.

The term \textit{sentiment analysis} lacks a unified definition. Generally, it refers to the automatic detection of a user's evaluative or emotional attitude toward a topic, as it is expressed in a text. More commonly, the meaning of \textit{sentiment} is restricted to the polarity (or valence) of a text, i.e. whether the text is positive, negative, or neutral \citep{mohammad2015sentiment}.

When a service or a product are the target of customer reviews, blog posts, or tweets, it may be sufficiently informative to determine the valence of the text. However, in many cases, expanding such binary---or ternary---categories to a set of chosen emotions yields higher explanatory power. From this claim rises the field of emotion classification, or emotion analysis, or non-binary sentiment analysis.

As in the case of binary sentiment analysis, there exist two main approaches to automatically extract affectual orientation: lexicon-based and corpus-based. The lexicon-based approach considers the orientation of single words and phrases in the document and it requires dictionaries of words labeled with the emotion---or emotions---they evoke. The corpus-based approach can be seen as a supervised classification task. Hence, it requires emotion-annotated corpora.
The performance of a statistical emotion classifier is typically good in the domain the classifier is trained on, but it can be mediocre when applied to other domains. Such classifiers lack generalizing power because their only source of information is the corpus they learn from, i.e. they are context-dependent. Consequently, the lexicon-based approach is often preferred as it provides higher context-independence \citep{taboada2011lexicon}.
An important limitation to the lexicon-based method is still the small size of the available lexical resources, which has a positive effect on precision at the cost of low recall. We propose a new lexicon expansion routine that addresses this shortcoming.

In the proposed framework, emotion-specific word embeddings are learned from a corpus of texts labeled with six basic emotions (anger, disgust, fear, joy, sadness, and surprise).
The derived vector space model is used to expand an existing emotion lexicon via a semi-supervised label propagation algorithm.

This thesis has multiple contributions. It introduces a novel variant of the Label Propagation algorithm that is tailored to distributed word representations. It applies batch gradient descent to accelerate the optimization of label propagation and to make the optimization feasible for large graphs. It proposes a reproducible method for emotion lexicon expansion, which can be leveraged to improve the accuracy of an emotion classifier.

An emotion-labeled corpus and an emotion lexicon are the two necessary resources for our method. As for the former, we use the Hashtag Emotion Corpus \citep{mohammad2015hashtag},
a collection of tweets
labeled with Ekman's six basic emotions \citep{ekman1992}.
The lexical resource we employ is the NRC Emotion Lexicon \citep{mohammad2013lexicon}.



The first step of our approach is to use the corpus as source of supervision for a deep neural network model. In particular, the core architecture is a Long Short Term Memory (LSTM) recurrent network \citep{hochreiter1997lstm}. The deep model learns emotion-specific representations of words via backpropagation, where the \textit{emotion-specificity} of a word vector refers to the ability to encode affectual orientation and strength in a subset of its dimensions.
Next, the specialized embeddings are employed to build a semantic-similarity graph. The emotion lexicon is expanded using our novel variation of the Label Propagation algorithm \citep{zhu2002labelprop}.

The rest of this thesis is structured as follows: we begin with a review of related work in the areas of emotion classification and lexicon expansion, and with a description of the statistical approaches that can be used to learn task-specific continuous representations (Section \ref{sec:relwork}). Then, the proposed lexicon expansion method and the optimization of specialized word embeddings are presented (Section \ref{sec:methods}). Section \ref{sec:analysis} presents an analysis of the employed resources. The experiments performed to learn emotion-specific embeddings and to expand the lexicon are reported in Section \ref{sec:exp} along with intrinsic and extrinsic evaluation in an emotion classification task (Section \ref{sec:eval}). Section \ref{sec:concl} concludes and proposes new research ideas.

The software related to this paper is open-source and available at \url{https://github.com/Procope/emo2vec}.

\section{Related work} \label{sec:relwork}

\subsection{Emotion classification} \label{relwork:emoclass}
Among the various areas of opinion mining, sentiment analysis is probably the most thoroughly explored. While sentiment analysis refers to the automatic detection of a user's affectual attitude toward a topic, more commonly, the goal of the field is to determine the \textit{polarity} of phrases, sentences, and documents (\textit{polarity} or \textit{valence annotation}). Hence, text can be typically classified as positive, negative, or neutral. It is, however, common to extend the number of valence classes to at least five: this finer-grained analysis additionally includes \textit{very positive} and \textit{very negative} in the polarity scale.

While sentiment analysis is an active and useful research area, it can only prove limitedly informative as it essentially maps text to a one-dimensional space. Indeed the usual range of valence is $[-1, 1]$.
Another possible goal of sentiment analysis is to still automatically detect a user's affectual orientation, but with the difference that text is assigned to one or more classes of emotions.
By increasing the number of dimensions used to represent emotional orientation, such opinion mining methods gain in interpretative and explanatory power. We refer to this extended problem as \textit{emotion classification} or \textit{emotion analysis}.
There are two possibility for the classification of texts into emotion categories: multinomial classification, where the classifier outputs a probability distribution over all emotions, and multi-label classification, where the classifier returns a probability for each emotion.

An important milestone for emotion analysis was the SemEval-2007 Affective Text task. The motivation for the task was that there seems to be a connection between lexical semantics and the way we verbally express emotions \citep{strapparava2007semeval}. In particular, it was argued that emotional orientation and strength of a text are determined potentially by all words that compose it, though, disputedly, in uneven amount. Therefore, expressions that appear to be neutral can also convey affective meaning as they might be semantically related to emotional concepts.
Consider the following tweets:
\ex. I want cake. I bet we don't have any. \label{tweet:cake}

\ex. Saddened by the terrifying events in Virginia. \label{tweet:timcook}

\ref{tweet:cake} conveys frustration, which, in terms of basic emotions, could be translated into the labels \textit{anger} and / or \textit{sadness} although its constituents appear to be neutral.
Similarly, \ref{tweet:timcook} clearly expresses an affectual orientation but the NRC Emotion Lexicon does not contain \textit{sadden}, \textit{saddened} nor \textit{terrify}, \textit{terrifying}.
The claim that all words potentially convey affective meaning is inspirational for our work and it provides a rationale for lexicon expansion (Section \ref{relwork:expansion} and \ref{meth:expansion}): since possibly all terms in a document contribute to its affective content, employing lexica of, at best, 15,000 types is a serious limitation.

The lexicon-based approach relies on labeled dictionaries to calculate the emotional orientation of a text based on the words and phrases that constitute it \citep{turney2002}.
A disadvantage of dictionaries is that they contain direct affective words, i.e. words that refer directly to affective states. In contrast, indirect affective words only have a weak connection to emotional concepts that depends on the context they appear in \citep{strapparava2006}. To give an example, an American professional baseball player, who was criticized for his unsatisfactory performance, publicly stated:
\ex. I am going to have a monster year.

The indirect affective word \textit{monster} is clearly used as a positive modifier, but context is required to make such an inference.

Finally, consider the following headline whose emotional orientation is rather positive.
\vspace{-0.4cm} \ex. Beating poverty in a small way. \label{headline}

 It contains the direct affective words \textit{beating} and \textit{poverty}, which are labeled as expressions of \textit{anger}, \textit{disgust}, \textit{fear}, and \textit{sadness} in the NRC Emotion Lexicon (Section \ref{relwork:resources}). Sentence \ref{headline} is an example of how lexicon-based methods cannot correctly analyze compositionality (\emph{beating poverty}). Negations and sarcasm are other such issues, which can only be solved adding rules, chunking, or parsing.

The use of dictionaries is not the only approach to emotion analysis. As it happens, there is another relevant method to address this task. The automatic extraction of affectual orientation can be viewed as a supervised classification problem, which requires an emotion-annotated data set and a statistical learning algorithm. This is often referred to as the \textit{corpus-based} approach \citep{pang2002}.

In the area of corpus-based methods, researchers have proposed different systems that, having access to contextual information, have the potential to address the issues of lexicon-based techniques: basic compositionality, negation, and cases of obvious sarcasm.

Two corpus-based systems participated in the SemEval-2007 Affective Text task: UA and SWAT.
UA uses Weighted Pointwise Mutual Information (PMI) to compute emotion scores. The distribution of emotions and the distribution of nouns, verbs, adjectives, and adverbs are obtained via statistical analysis of texts crawled from Web search engines. Then, given the set of words $W_t$ in a text $t$, and an emotion word $e$, PMI is defined as: $$PMI\left(t, e\right) = \frac{fr\left(W_t \cap \{e\} \right)}{fr\left( W_t \right) fr \left( e \right)}$$  Additionally, an associative score between content words and emotions is estimated to weight the PMI score.

An even more traditional supervised system is SWAT. Based on a unigram model, SWAT additionally uses a thesaurus to obtain the synonyms of emotion label words (e.g. the synonym of \textit{joy}, or \textit{surprise}).
Finally, \citet{strapparava2008own} proposed a Na\"ive Bayes classifier trained on mood-annotated blog posts.

While UA and SWAT were the corpus-based systems participating in the SemEval-2007 Affective Text task, UPAR7 is an example of the lexicon-based counterpart. UPAR7 can be described as a linguistic approach. It parses a document and uses the resulting dependency graph to reconstruct what is said about the main subject. All words in the document have emotion scores based on SentiWordNet \citep{esuli2007sentiwordnet} and WordNet Affect \citep{strapparava2004wordnet}, enriched with lexical contrast and accentuation. A higher weight is given to the score of the main subject, while the other weights are computed considering linguistic categories such as negation and modal verbs.

There exist as well combinations of the two main approaches. \citet{strapparava2008own} used, on the one hand, lexical resources (WordNet Affect) to annotate synsets representing emotions and moods. For each emotion, a list of words is generated by the corresponding synset. On the other hand, the British National Corpus is fed into a variation of Latent Semantic Analysis (LSA) to obtain a vector space model, where words, documents, and synsets are represented as vectors. Documents and synsets are mapped into the vector space by computing the sum over the normalized LSA vectors of all the words comprised in them.
Once an emotion is represented in the LSA vector space, determining affective orientation is essentially a matter of computing a similarity measure between an input word, paragraph, or text, and the prototypical emotion vectors.
This was the best performing system at SemEval-2007, which suggests that lexical resources and corpora are complementary sources of information.

\citet{yang2015lcct} combined the two main learning methods in a unified co-training framework applied to valence annotation, obtaining state-of-the-art results on various English and Chinese datasets.
Indeed, further studies \citep{kennedy2006, andreevskaia2008, qiu2009} showed that the performance of lexicon- and corpus-based approaches is complementary in terms of precision and recall. The lexicon-based method yields higher recall at the cost of low precision, as it is likely to tag an instance with label \textit{l} whenever an emotion word related to \textit{l} is found---i.e. even if the text is neutral or rather evocative of another emotion.  In contrast, supervised learning systems tend to perform poorly in terms of recall---since their vocabulary is limited to the types seen in the training data---but they reach higher precision scores as labels are assigned in a more fine-grained manner \citep{yang2015lcct}---after all, most training instances have a considerable neutral content (Section \ref{sec:analysis}).

None of the systems participating to SemEval-2007 reached an accuracy similar to that of sentiment analysis methods. This is possibly the reason why it is hard to find literature about alternative approaches to those used in the Affective Text task, and why a similar task was not proposed in successive editions of SemEval. However, the inter-annotator agreement studies conducted by \citep{strapparava2007semeval} for the SemEval headlines corpus, remind us that there is an upper bound to the accuracy of emotion classification algorithms, namely human accuracy. Table \ref{table:interann} shows the agreement scores in terms of Pearson correlation, as they can be found in the original paper. If trained annotators agree with a simple average \textit{r} of 53.67 (the frequency-based average \textit{r} is 43), it is plausible that untrained respondents would show an even lower level of agreement. In Section \ref{sec:eval-classification} we describe the results of a survey that tests the ability of an average untrained person to classify short paragraphs into emotions.

A comprehensive list of challenges in the analysis of affective text is presented by \citet{mohammad2015sentiment}, and it includes e.g. subjective and cross-cultural differences. In the light of these considerations, NLP researchers should be encouraged to consider emotion classification as a field with potential. Low accuracy scores are explained by a low upper bound.

\begin{table}[t]
    \centering
     \begin{tabular}{c c c c c c}
        anger & disgust & fear & joy & sadness & surprise\\
        \hline
        49.55 & 44.51 & 63.81 & 59.91 & 68.19 & 36.07\\[1ex]
    \end{tabular}
    \begin{tabular}{c c}
    simple average & frequency-based average \\
    \hline
    53.67 & 43
    \end{tabular}
    \caption{Pearson correlation for inter-annotator agreement on the SemEval-2007 Affective Text Corpus \citep{strapparava2007semeval}.}
    \label{table:interann}
\end{table}

\subsection{Available corpora and lexica} \label{relwork:resources}
An important attempt to organize affective phenomena was made by \citet{ekman1992}, who introduced the concept of basic emotions, i.e. affective states that seem to share a connection with physiological processes and universal facial expressions. Whether it is possible to identify a fixed number of categories, out of the many moods, attitudes, and traits, is outside the scope of this thesis. However, since we deal with emotion classification, it is relevant to have a set of meaningful classes at our disposal.

While some researchers disagree on the validity of the theory of basic emotions, others have different opinions as to which affective states constitute a justifiable set of elementary categories. \textit{Ekman's Six} include anger, disgust, fear, joy, sadness, and surprise. Adding trust and anticipation yields Plutchik's eight \textit{primary} emotions \citep{plutchik1980}. A collection of seven categories was used in the ISEAR project: joy, fear, anger, sadness, disgust, shame, and guilt \citep{scherer1994}, whereas \citet{izard1971} counted nine basic emotions.

While it is possible to choose from a variety of sets of affective states, the scarce amount of annotated datasets constrains the choice of researchers.
The SemEval-2007 Affective Text corpus is a collection of news titles extracted from newspapers and news web sites \citep{strapparava2007semeval}. It consists of 1250 headlines labeled with Ekman's six emotions. In an attempt to obtain fine-grained labels, six annotators independently assigned every headline--emotion pair an emotion score ranging $[0,100]$, which indicates how intensively a news title conveys the corresponding emotion.

The data was made available in two sets: a development set consisting of 250 headlines, and a test set of 1,000 headlines. This partition reflects the fact that the SemEval-2007 Affective Text task was aimed at unsupervised approaches. In later work on supervised classification methods, the 1,000 news titles were used as training data, and the remaining 250 for testing \citep{mohammad2015hashtag}.
Moreover, in many experiments, the vector of six emotion scores was made coarser-grained in order to better fit classification algorithms. In single-label settings, only the most dominant emotion, i.e. the emotion with the highest score, was used as headline label \citep{chaffar2011}. For multi-label classification, only emotions with a score higher than a threshold \textit{k} were considered present in a given headline. While the task description of SemEval-2007 Affective Text indicates $k=50$, other researchers set the threshold to a lower value (e.g. $k=25$ was used by \citet{mohammad2015hashtag}).

The SemEval dataset does not contain text from social media platforms, which are nowadays the most common target of opinion mining. An attempt to cover this type of content was made by \citet{mohammad2015hashtag}, who created a corpus of tweets (Twitter posts) leveraging the use of \textit{hashtags} (i.e. words immediately preceded by a hash symbol which mostly serve to signal the topic of Twitter posts, as well as other sorts of metadata such as the tweeter's mood). The Hashtag Emotion Corpus consists of about 21,000 tweets annotated with one out of the six emotions proposed by Ekman.

Other corpora were compiled with sentence-level annotation. Sentences extracted from 22 fairy tales were annotated by \citet{alm2005} with 5 emotions (joy, fear, sadness, surprise, and anger-disgust). \citet{aman2007} annotated about 5,000 sentences drawn from blog posts with Ekman's six emotions and a \textit{neutral} category. \citet{neviarouskaya2009} chose the nine emotion categories proposed by Izard (and a \textit{neutral} one) to label 1,000 sentences extracted from stories on a variety of topics.

Lexical resources are not abundant either. WordNet Affect is a word-emotion association lexicon consisting of 1,536 terms \citep{strapparava2004wordnet}. In occasion of the SemEval Affective Text task, a set of 152 words extracted from WordNet Affect was made available for optional use. The Hashtag Emotion Lexicon contains 11,418 lemmas automatically obtained from the Hashtag Emotion Corpus \citep{mohammad2015hashtag}. Each word-emotion pair comes with a real-valued association score, namely the Strength of Association (SoA) score between a word $w$ and emotion $e$: $SoA(w,e) = PMI(w,e) - PMI(w,\lnot e)$. While the three above lexica use Ekman's Six, the NRC Emotion Lexicon \citep{mohammad2013lexicon} combines the eight primary emotions proposed by Plutchik with the positive and negative polarities, for a total of ten possible labels. The NRC Emotion Lexicon includes 14,182 English words. It was manually created via crowdsourcing and uses binary word-emotion association scores.

\subsection{Learning task-specific continuous representations} \label{relwork:taskspec}
Supervised methods of statistical learning require input data to be represented by features. In NLP there are many ways to encode words, sentences, and paragraphs. While it was common to hand-engineer linguistically motivated features,
the field now prefers automatic feature generation as it does not demand domain expertise nor extensive manual work, and it allows to define language-independent techniques and models.

A simple yet common way to represent words are
\textit{one-hot} vectors, which can, in general, be used to encode categorical features. Each word is represented as a boolean vector whose length is equal to the size of the vocabulary: if a word is at position $i$ in the vocabulary, only the $i$th dimension of the word vector is set to 1.
As a result, words are represented by large, sparse vectors which can be fed into various statistical learning algorithms, such as linear classifiers.
Although the one-hot representation can be improved by using e.g. counts instead of boolean values, it can encode virtually no information about the relation between words and their similarities.

As deep learning started to regain popularity, alternative encodings resurged. Sparse vectors were largely replaced by dense continuous representations, \textit{word embeddings}, which have proved effective in multiple NLP tasks such as parsing, language modeling, named entity recognition, machine translation, and word sense disambiguation.

Particularly interesting for our research is the field of statistical language modeling, which successfully uses distributed representations of words as a paradigm \citep{rumelhart1988}. Unlike one-hot vectors, word-embeddings do not represent words in isolation. They rather exploit the context of a word to learn its syntactic and semantic properties.

\citet{mikolov2013w2v} proposed two architectures for efficiently estimating vector representations of words from large corpora: the Continuous Bag-of-Words model (CBOW) predicts a word based on its context, whereas the Skip-gram model predicts a context given a word.
CBOW takes as an input the context of a target word, i.e. the window consisting of the $n$ history words and the $n$ future words, where $n$ is the context window size. As any bag-of-words model, CBOW ignores the order of context words. Its novelty lies in the use of continuous vectors to represent the context.
The Skip-gram model operates in the opposite direction. Receiving a word as an input, it predicts the most likely previous and next words. Similarly to CBOW, Skip-gram allows variable-size context windows. Skip-gram intuitively formulates the probability of a context word $w_c$ given an input word $w_i$ in terms of the softmax function:
\[
P\left( w_c | w_i \right) = \frac{exp\left( v_{w_c}^{\ \ T} \  v_{w_i} \right)}{\sum_{w=1}^{|V|} exp\left( v_w^{\ \ T} \ v_{w_i} \right)}
\]
However, the computation cost for the derivative of each probability is proportional to $|V|$, the number of words in the vocabulary.
An alternative way of calculating the probability of a context word is Negative Sampling, a simplification of Noise Contrastive Estimation \citep{gutmann2012noise}. The task is then to distinguish a candidate context word from instances drawn from the noise distribution of the input word. Negative Sampling is preferred as it significantly reduces computation costs.

CBOW and Skip-gram are prominent examples of state-of-the art models. Models such as those proposed by \citet{collobert2011scratch} and \citet{pennington2014glove} are also frequently used in many different NLP tasks.

Word representations may vary depending on the model they originated from but, most interestingly, they may vary according to the task they are applied to. For example, neural language models can be specialized to learn the syntactic properties of words, such as their order of occurrence and simple collocations, which can be useful for part-of-speech tagging and parsing. Typically, a small-sized context is used for this goal.  As another option, a model can be specifically trained to learn the word semantics using larger contexts and it will perform well on analogy questions such as man:king = woman:\textit{x} (as in ``man is to king as woman is to \textit{x}'').
The resulting word embeddings show intriguing properties. First, words with similar meaning are mapped to a similar position in the vector space. For example, \emph{school} and \emph{university} are represented by similar vectors, i.e. they are close to each other in the vector space, while \emph{school} and \emph{green} are more distant \citep{mikolov2013distributed}. Furthermore, it is possible to do linear vector calculations to answer syntactic and semantic analogy questions. Analogies such as the one introduced above can be answered by computing $x = x_{king} - x_{man} + x_{woman}$ \citep{mikolov2013regularities}. On a syntactic level, it was observed that models can learn the offset corresponding to the concept of plurality: e.g. $x_{apple} - x_{apples} \approx x_{car} - x_{cars}$.  Similar results were obtained by \citet{turney2012} using a relational similarity model, and by Glove vectors \citep{pennington2014glove}. In fact, the Glove vectors outperform both Skip-gram and CBOW on many analogy tasks.
Finally, \citet{mikolov2013translation} showed that distributed representations can be used to translate words between languages by learning a linear transformation matrix that maps embeddings of the input language to the \textit{translated} embeddings in the same vector space.

In the field of sentiment analysis, it is possible to learn sentiment-specific word embeddings directly from a large annotated corpus. \citet{tang2014specific} introduced such a method by extending an existing \textit{general purpose} embedding algorithm, the Collobert and Weston (C\&W) model \citep{collobert2011scratch}. The resulting representations show a better predictive performance in a supervised sentiment analysis task.

C\&W belongs to the family of models that predict a word based on context. In particular, given an ngram $t$ ``the boy with a telescope", a corrupted ngram $t_r$ is derived by substituting the center word with a random word $w^r$: ``the boy $w^r$ a telescope". The training objective is that the original ngram obtains a higher language model score $f_{cw}\left(t\right)$ than its corrupted version by a margin of 1:
\begin{align}
	loss_{cw} \left( t, t_r \right) = max \left(  0, 1 - f^{cw}\left( t \right) + f_{cw} \left( t_r \right) \right)
\label{eq:cw_loss}
\end{align}

The architecture of the C\&W neural model consists of a lookup layer with a table $L$, a linear layer with parameters $w_1, b_1$, a non-linear  $hTanh$ layer, and a second linear layer with parameters $w_2, b_2$. The language model score for an ngram is then defined as:
\[f_{cw}\left( t \right) = w_2(a) + b_2 \]
\[a = hTanh\left(w_1 L_t +  b_1 \right) \]
	\[
	hTanh(x)=\left\{
							\begin{array}{ll}
								-1 \ &\textrm{if} \ \ x < -1\\
								x &\textrm{if} \ -1 \leq x \leq 1\\
								1 &\textrm{if} \ \ x > 1
							\end{array}
					\right.
	\] \smallskip

\citet{tang2014specific} integrate the C\&W model by learning an additional function $\hat{s}\left( t \right)$ which predicts the sentiment score of ngrams. The sentiment-specific loss is defined as:
\begin{align}
	loss_s \left( t, t_r \right) = max \left(     0 , 1 - \delta_s \left( t \right) \hat{s}_0 \left( t \right)
																	  +  \delta_s \left( t_r \right) \hat{s}_1 \left( t \right)
											    \right)
\label{eq:sent_loss}
\end{align}
where $s\left( t \right)$ is the gold sentiment distribution of an ngram, and $\delta_s$ is an indicator function:
\[
	\delta_s \left( t \right) = \left\{
												\begin{array}{ll}
														  1 \ &\textrm{if} \ \ s\left( t \right) = \left[ 1 , 0 \right] \\
														-1 \ &\textrm{if} \ \ s\left( t \right) = \left[ 0 , 1 \right]
	\end{array}
	\right.
\]
There are now two concurrent objectives: modeling the syntactic context of words ignoring sentiment, and learning the polarity of words based on the affective context rather than the syntactic one.
The two models can be made complementary by computing the linear combination of the hinge losses in Equations \ref{eq:cw_loss} and \ref{eq:sent_loss}, weighted by a hyperparameter $\alpha$:
\begin{align}
loss_s \left( t, t_r \right) = \alpha \ loss_{cw}\left( t, t_r \right) \ + \ \left(1 - \alpha \right) loss_{s}\left( t, t_r \right)
\end{align}

This extension of the C\&W model produces embeddings which encode polarity information \citep{tang2014specific}. Hence, such representations constitute useful features for polarity annotation tasks. It must be noted, however, that a very large annotated corpus was used to train the model. Indeed, it was shown that the quality of the specialized vectors is directly proportional to the size of the data set. In a polarity classification task, the model trained on 1 million tweets yields an $F_1$ score of 0.78, whereas the model trained on 12 million tweets obtains an $F_1$ score of 0.83.

All the language models described above produce good quality word embeddings when they are trained on corpora of very large size, in the $10^6 - 10^9$ range. The field of emotion classification suffers from the lack of very large annotated corpora.  To our knowledge, the largest available data set is the Hashtag Emotion Corpus, which contains about 21,000 tweets. This size range is likely to be insufficient to learn a language model from scratch, therefore we need a technique that is able to learn task-specific distributed representations even from smaller data sets. \citet{labutov2013} proposed a method that takes as input pretrained word embeddings and some labeled data, and rearranges the embeddings in their original vector space, without directly learning an entire task-specific language model. This alternative approach has multiple advantages: the task-specialization process of word embeddings is computationally more efficient, the size of the corpus need not be overly large, pretrained generic word embeddings can be leveraged, and an established model can be used to learn them.

\subsection{Lexicon expansion and semi-supervised learning} \label{relwork:expansion}
The task of expanding a lexicon can be solved using a variety of methods. The most na\"ive approach would be to use \textit{self-training}, i.e. first labeling a portion of the unlabeled data based on the few labeled instances, and then using the newly labeled data points to incrementally classify the whole unlabeled portion of the tokens. However, since dictionaries are typically not very large, the initial classifier performs poorly as it cannot exploit the great amount of information provided by the relations between unlabeled words, which constitute the vast majority of the available data.
As a consequence, the accuracy of a self-training classifier decrements at each iteration.

A different method is \textit{transductive inference} \citep{vapnik1998statistical}. In transductive learning, a learner $L$ is given a hypothesis space $H = \left\lbrace h\ \mid \ h : X \to \{-1,1\} \right\rbrace$, a training dataset $D_{train}$ and a test dataset $D_{test}$ from the same distribution. The learner then tries to find the function $h_L = L(D_{train}, D_{test})$ that minimizes the expected number of erroneously classified instances of the test set \citep{joachims1999transductive}.

Transductive SVMs (TSVMs) have been successfully used for text classification but they have a crucial pitfall: the TSVM optimization problem is combinatorial. Although \citet{joachims1999transductive} proposed an algorithm that finds an approximative solution using local search, in order to keep the optimization problem tractable, the size of test sets cannot exceed 10,000--15,000 instances. If this threshold is probably too low compared e.g. to the about 30,000 word types that compose the Hashtag Emotion Corpus, it is surely intolerable if we use, as a benchmark, the more than 300,000 word types with frequency $fr \geq 150$ of the ENCOW14 corpus \cite{schafer2012cow,schafer2015cow}. An additional downside is that the TSVM expects sparse input vectors, hence it disallows the use of dense word embeddings.

A more linguistically inclined approach is to compute the \textit{semantic orientation} of words based on the PMI between tokens and emotion words---or, on Twitter, emoticons. This method has been used to expand a lexicon for context-dependent polarity annotation \citep{zhou2014}.

The problem of lexicon expansion can also be conceived as a supervised classification task, where the words in the dictionary are used for training. \citet{bravo2016expansion} recently deployed a corpus of ten million tweets \citep{petrovic2010edinburgh} and a multi-label classifier to expand the NRC Emotion Lexicon. The proposed classifiers are of three types: Binary Relevance, Classifier Chains, and Bayesian Classifier Chains. They all use word-level features, which can be extracted with the Skip-gram model, or the word-centroid model. Although the latter draws information from multiple features---word unigrams, Brown clusters, POS ngrams, and Distant Polarity--- word embeddings learned via the Skip-gram model were shown to significantly outperform word-centroid features at boosting classification performance.

The disproportion between lexicon words and unseen types signals, however, that a semi-supervised learning technique appears to more naturally fit the expansion problem.

From the perspective of a distributional semanticist, words float in a high-dimensional space. This configuration is suitable for building a graph, where words are regarded as nodes linked by weighted edges. Representing the semantic space as a graph is particularly useful because graphs are very tractable mathematical objects, which come with a large variety of optimized algorithms.

Graph-based and semi-supervised, the Label Propagation algorithm seems to represent a strong alternative to multi-label classification.
This is an iterative algorithm that propagates labels from labeled to unlabeled data by finding high density areas \citep{zhu2002labelprop}. All words, labeled and unlabeled, are defined as nodes. The edge between two nodes $w_1$, $w_2$ is weighted by a function of the proximity of  $w_1$ and $w_2$, i.e. words that are close in the semantic space are linked by strong edges.
Moreover, all nodes are assigned a probability distribution over labels. As we iterate, labels propagate through the graph and the probability mass is redistributed following a crucial principle: labels propagate faster through strongly weighted edges.
Although Label Propagation requires hyperparameter optimization, it can efficiently solve the label propagation problem without iteration, as it was shown to have a unique solution.

This technique appears to be the most appropriate for lexicon expansion as it leverages dense word vectors and their semantic similarities. Moreover, since word embeddings can be learned from corpora, they carry the context-dependent information that a purely lexicon-based classifier typically lacks.

\section{Methods} \label{sec:methods}
\subsection{Emotion-specific embeddings} \label{meth:emospec}

\subsubsection*{Task specificity}
Neural language models such as Skip-gram or CBOW are based on the fundamental idea that an unsupervised problem can be solved by embedding it in a supervised task. In particular, neural language models predict a word given a context or a context given a word in a supervised manner. Each word is represented as a vector that is free to vary to improve the performance of the supervised task. As a result, word embeddings are specialized in representing the relation between a word and its habitual context. This relation can be considered to stand for the concept itself, thus the obtained representations are optimal word features in a variety of tasks.

In a similar attempt, we learn task-specific word embeddings via a supervised task. Since we are interested in embeddings that encode affective content, our supervised problem is emotion classification. Beforehand, the CBOW model is used to learn general purpose word representations from a large unsupervised corpus, ENCOW14. These will serve as an informed initialization for the model weights.

\subsubsection*{Motivation for recurrent neural networks}
Traditional neural networks are not able to make full use of sequential information, as they act under the assumption that all inputs occur independently of each other. In contrast, (written) natural language is sequential in its nature as paragraphs and sentences are not successions of words randomly drawn from a vocabulary. The use of a specific word is dependent on the word's context.

Recurrent Neural Networks (RNNs) address the sequentiality issue as they include an artificial memory that allows them to make decisions based on past time steps. Thus, RNNs have the ability to represent context. In practice, information persists due to the structure of recurrent networks: they consist of multiple copies of the same network applied to different time steps.

Nonetheless, not all types of RNNs are appropriate for the analysis of natural language. As a case in point, the Simple Recurrent Network (SRN) introduced by \citet{elman1990} cannot properly handle a typical natural language phenomenon, non-local dependencies. A syntactic dependency is considered non-local when it involves sentence constituents that are ``out of place'', i.e. their function cannot be simply derived by their position in the sentence. For example, consider that \ref{nonloc-a} can be paraphrased as \ref{nonloc-b}.
\ex.
    \a. John is precisely looking for this book. \label{nonloc-a}
    \b. It is precisely this book that John is looking for. \label{nonloc-b}

In this case, the phrase \textit{this book} does no longer follow the verb it is syntactically related to.

Linguists have identified many analogous phenomena, such as topicalization and wh-movement. Particularly difficult to handle are unbounded dependencies, where a phrase moves to an arbitrarily long distance from its usual position. \ref{nonloc-b} can be modified to exemplify a long-distance dependency:
\ex. It is precisely this book Matthew said Mark believes Luke knows that John is looking for. \label{longdist}

This example is deliberately exaggerated in an attempt to demonstrate the \textit{unboundedness} of some non-local dependencies. Nevertheless, less extreme long-distance dependencies are used in many languages and deserve appropriate treatment if we aim e.g. for a good model of English.

It should be noted that SRNs are, in theory, able to exploit distant information. However, they fail due to the properties of gradient-based learning and backpropagation. Neural network weights receive updates proportional to the gradient of the loss function, which are computed via the chain rule in multi-layer architectures. Repeatedly multiplying gradients has the effect of either increasing or decreasing the error exponentially. These behaviors are known respectively as the exploding gradient problem, and the vanishing gradient problem. At first sight, Simple Recurrent Networks do not seem to be affected by exploding or vanishing gradients as they are---at least in their basic formulation---one-hidden-layer architectures. A more careful examination reveals that backpropagation through time \citep{werbos1990bptt} actually unfolds the recurrent network into a series of feedforward layers, exposing the remote time steps to exploding and vanishing gradients. As a consequence, SRNs are too sensitive to recent context and indifferent to remote time steps \citep{hochreiter1991, bengio1994learning}.

Consider Elman network as an example:
\begin{gather*}
h_t = \sigma _{h} \left(W_{h}\ x_{t}+U_{h}\ h_{t-1}+b_{h}\right)\\
y_{t} =\sigma _{y}\left(W_{y}h_{t}+b_{y}\right)
\end{gather*}
The hidden layer of the current time step $t$ is a weighting of the hidden layer of the previous time step and the input of the current time step, to which a non-linearity is applied.
In the case of a long-distance dependency, the gradient should backpropagate through multiple time steps.
However, (i) many non-linearities, such as the logistic function and the hyperbolic tangent function, have small gradients in the tails and (ii) these small gradients are multiplied due to the chain rule of differentiation. Hence the resulting product tends to exponentially decrease as a function of the number of previous time steps. This is the reason why long-distance dependencies between time steps are difficult to learn.

A solution is fortunately provided by Long Short-Term Memories (LSTMs), a variant of recurrent neural networks that is designed to overcome the exploding and vanishing gradient problem by enforcing constant error flow \citep{hochreiter1997lstm}.
More precisely, LSTMs address the unstable gradient problem by propagating information between time steps using a vector---the cell state---that is a linear combination of the previous cell state and the new candidate hidden state. This allows the gradient to flow through time, making it possible to capture long-distance dependencies.

There remains, nevertheless, one challenge: LSTMs only consider the left context of an input. Since we deal with natural language, where sentences have a non-linear structure, access to the right context of a word also provides relevant information. As an example, consider this tweet from the :
\ex. My niece calling to sing Happy Birthday to me \#love !! \label{tweet}

If one wants the affective orientation of \textit{love} to percolate to \textit{Happy Birthday}---or perhaps even to \textit{niece}---, access to backward-flowing information is necessary.
Such a bidirectional information flow can be obtained by using two recurrent networks that are presented each sequence forwards and backwards respectively. Connected to the same output layer, the two networks provide complete sequential information about every time step \citep{graves2005bidir}. This property motivates our use of a bidirectional LSTM.

\subsubsection*{Model}
The inputs to our emotion classifier are paragraphs. An embedding layer maps words to their vector representations. The embeddings are fed to a bidirectional LSTM, followed either by a softmax layer that outputs probability distributions over emotion classes or by a sigmoid layer that produces one probability value for each class.
Batch normalization precedes both the bidirectional LSTM layer and the output layer.

The forward and backward LSTMs, which receive an input sequence $x_0, x_1, \ldots, x_n$ and output a representation sequence $h_0, h_1, \ldots, h_n$, are implemented as follows.
To compute the new state of a memory cell at time $t$ we need the value for the input gate $i_t$, the candidate value $\widetilde{C}_t$ for the cell state, and the activation of the forget gate.
\begin{gather}
\label{eq:lstm_first}
i_t = \sigma(W_i \ x_t + U_i \ h_{t-1} + b_i)\\
\widetilde{C}_t = tanh(W_c \ x_t + U_c \ h_{t-1} + b_c)\\
f_t = \sigma(W_f \ x_t + U_f \ h_{t-1} + b_f)
\end{gather}
The new state $C_t$ of the memory cell is then computed as
\begin{align}
C_t = i_t * \widetilde{C}_t + f_t * C_{t-1}
\end{align}
Finally, we compute the activation of a cell's ouput gate and then the cells's output:
\begin{gather}
o_t = \sigma(W_o \ x_t + U_o \ h_{t-1} + b_o) \\
h_t = o_t * tanh(C_t)
\label{eq:lstm_last}
\end{gather}
In Equations \ref{eq:lstm_first}-\ref{eq:lstm_last}, $W_i, W_f, W_c, W_o, U_i, U_f, U_c$, and $U_o$ are independent weight matrices, while $b_i, b_f, b_c$, and $b_o$ are independent bias vectors.

The use of batch normalization is motivated by its ability to improve training speed by allowing higher learning rates, to reduce the importance of a careful initialization, and to possibly act as a regularizer \citep{ioffe2015batch}.
This technique was introduced to reduce covariate shift, a known neural network problem: as the network's parameters are updated during training, the distribution of the activation also changes. Since networks converge faster if their inputs have zero means and unit variances \citep{lecun2012efficient}, batch normalization attempts to fix the distribution of the inputs to any network layer, producing a significant speedup in training.

For a $d$-dimensional input $x = \left( x^1 \ldots x^d \right)$, each dimension $k$ is normalized as follows:
\begin{align*}
\widehat{x}^k = \frac{x^k - E\left( x^k \right)}{\sqrt{Var \left( x^k \right)}}
\end{align*}
As we use batch-based stochastic gradient descent training, the expectation $E\left( x^k \right)$ and the variance $Var \left( x^k \right)$ can be estimated for each batch. The output is then scaled and shifted by the learnable parameters $\gamma^k$, $\beta^k$, which conserve the network's representation power \citep{ioffe2015batch}.
Finally, the normalized output for batch $b$ is computed as:
\begin{align*}
y_i = \gamma \ \frac{x_i - \mu_b}{\sqrt{\sigma^2_b}} + \beta
\end{align*}
where $\mu_b$ is the batch mean, and $\sigma^2_b$ is the batch variance.
Since batch normalization is applied independently to each activation $x^k$, we omitted $k$ in the last equation.

Batch normalization was shown to act, in some networks, as a regularizer, as well as a method to increase training speed. In our model, we explore two other regularization techniques in order to avoid overfitting: $\ell_2$ regularization and dropout. L2 regularization consists of adding a penalty term $R\left( \theta \right) = ||\theta||^2_2$ to the objective function, where $\theta$ are the trainable model parameters. On the other hand, randomly removing units from a network during training is referred to as dropout \citep{srivastava2014dropout}.
In more detail, dropout consists of temporarily removing learning units and their connections from the network. Each unit has a probability $p$ of being dropped, which can be set as a hyperparameter. Since $p$ is a fixed, independent probability, each unit needs to learn to work with any randomly chosen subset of the network. That is, a learning unit cannot rely on other units, as it is not certain that those are present in the \textit{thinned} network. As a consequence, units are forced to learn only relevant features. This characteristic motivates the use of dropout.
We expect dropout to guarantee a better performance as it was specifically designed to prevent overfitting in neural networks.

\subsection{Lexicon expansion} \label{meth:expansion}
The lexicon expansion task is defined as follows. We are given a set of emotion classes $C$, and a set $W$ of word types extracted from a large corpus, which can be partitioned into $L \subset W$, the set of lexicon words, and $U \subset W$, the set of unlabeled words. For ease of notation, we refer to the set cardinalities as $|C| = m$, $|L| = l$, and $|U| = u$. Typically $l \ll u$.
We try to find a labeling function that maps each unlabeled word to a probability distribution over $m$ classes:
\begin{align*}
		\lambda : \ &U \to \mathbb{R} ^{m} \\
						 &w \mapsto (y_1, \ldots, y_m), \ \ s.t. \  \sum_{i=0}^{m}y_i = 1
\end{align*}

We choose the Label Propagation (LP) algorithm \citep{zhu2002labelprop} and propose a novel variant thereof in order to solve the lexicon expansion problem. LP is a graph-based semi-supervised technique that propagates labels from labeled to unlabeled nodes through weighted edges. Conceived as an iterative transductive algorithm, LP was shown to have a unique solution. It can therefore learn $\lambda$ directly, without iteration.

\subsubsection*{Problem setup}
More formally, let $\left(w_1, y_1 \right), \ldots, \left(w_l, y_l \right)$ be the labeled data $L$, and $Y_L = \{y_1, \ldots, y_{l}\}$ the label distributions thereof. $W$ is defined as a subset of $\mathbb{R}^d$, where $d$ is the number of dimensions used to encode words as continuous dense vectors. The goal is to estimate the label distribution of unlabeled data $Y_U$ from $W$ and $Y_L$.

To do so, we build a fully connected graph using labeled and unlabeled words as nodes. Edges between nodes are defined so that the closer two data points $x_i, x_j$ are in $d$-dimensional space, the larger the weight $w_{ij}$.

In the original version of Label Propagation, weights are defined in terms of a distance metric (euclidean distance) and they are controlled by a hyperparameter $\sigma$:
\begin{align}
w_{ij} = exp\left(-\frac{dist\left(x_i, x_j\right)^2}{\sigma^2}\right)
\end{align}
However, cosine similarity is commonly preferred to euclidean distance as a metric for word embeddings. We therefore define weights as:
\begin{align}\label{eq:weight}
w_{ij} = \sigma \left( \alpha \left(\frac{x_i \cdot x_j }{||x_i||_2 \ ||x_j||_2} \right) + b \right)
\end{align}
The use of the logistic function and the bias is motivated by the need (i) to adapt the weight computation to the properties of cosine similarity and (ii) to obtain a uniform weight formula regardless of the number of parameters (see \textit{Hyperparameters} subsection).

Further define a $\left( l+u \right) \times \left( l+u \right)$ probabilistic transition matrix T such that
\begin{align}
T_{ij} = P\left(i \to j\right) = \frac{w_{ij}}{\sum_{k=1}^{l+u}w_{kj}}
\end{align}
and a $\left( l+u \right) \times m$ label matrix $Y$, where $Y_i$ stores the probability distribution over labels for node $x_i$.
Notice that $T$ is column-normalized, so that the sum of the probabilities of moving to node $j$ from any node $i$ amounts to 1.

$Y_L$ rows are initialized according to the lexicon. If the lexicon only provides one label per word, then a probability value of $1$ will be assigned to the corresponding label. Since the NRC Emotion Lexicon maps words to multiple labels, we uniformly distribute the probability mass among all positive classes. For the initialization of $Y_U$, which \citet{zhu2002labelprop} consider not relevant, we assign a constant probability of $\frac{1}{m}$ to every label.

\subsubsection*{Algorithm}
First, the transition probability matrix $T$ needs to be row-normalized ($\bar{T}_{ij} = T_{ij} \ / \ \sum_{k} T_{ik}$), so that the sum of the probabilities of moving from node $i$ to any node $j$ amounts to 1.
Then, $\bar{T}$ is partitioned into 4 sub-matrices:
\begin{align}
	\bar{T} = \left[ \bar{T}_{ll} \ ; \  \bar{T}_{lu} \ ; \  \bar{T}_{ul} \ ; \ \bar{T}_{uu} \right]
\end{align}
The iterative algorithm essentially consists of the following update:
\begin{align}
	Y_U \leftarrow \bar{T}_{uu}Y_U + \bar{T}_{ul}Y_L
\end{align}
However, it was shown \citep{zhu2002labelprop} that the original algorithm converges to a unique solution:
\begin{align}
	Y_U = \left( I - \bar{T}_{uu} \right)^{-1} \bar{T}_{ul}Y_L
\end{align}

\subsubsection*{Hyperparameters}
 \citet{zhu2002labelprop} defined edge weights as $w = exp\left(-dist^2 / \sigma^2\right)$. 
On account of the properties of the exponential function, large sigmas increase edge weights, whereas small sigmas shrink them. In more precise words, when $\sigma \to 0$, the label of a node is mostly influenced by that of its nearest labeled node; when $\sigma \to \infty$, the label probability distribution of a node reflects class frequency in the data, as it is affected by virtually all labeled nodes in the graph.

\citet{zhu2002labelprop} presented two techniques to set the parameter $\sigma$. The simplest one is to find a minimum spanning tree (MST) over all nodes. Kruskal's algorithm \citep{kruskal1956mst} can be used to build a tree whose edges have the property of connecting separate graph components. The length $d_0$ of the first edge connecting components characterized by different labeled points is used as an approximation of the minimum distance between class regions. Finally, $\sigma = d_0 / 3$, so that the edge connecting two separate graph regions has a weight approaching $0$.

The second approach is to use gradient descent in order to find the parameter $\sigma$ that minimizes the entropy $H$ of the predictions.
\begin{align}
	H = - \sum_{ij}^{}\ Y_{ij} \ log Y_{ij}
\end{align}
This technique further allows the extension of one parameter $\sigma$ to $d$ parameters $\Sigma = \sigma_1, \ldots, \sigma_d$ that control edge weights along the $d$ dimensions used to encode each node. In our formulation of label propagation (Equation \ref{eq:weight}), $\Sigma$ translates into a vector $\vec{\alpha} \in \mathbb{R}^d$:
\begin{align}
    w_{ij} = \sigma \left(\vec{\alpha} \cdot \left( \frac{x_i}{||x_i||_2} \odot \frac{x_j}{||x_j||_2} \right) + b\right)
    \label{eq:weight_sigmas}
\end{align}
Each weight can be therefore interpreted as a cosine similarity, where every elementwise multiplication ($\odot$ is the Hadamard product) is scaled by a dimension-specific $\alpha$.
This formulation gives the algorithm the ability of discerning relevant dimensions, and the power to reduce the weight of irrelevant ones.

Gradient descent is used to find the parameters $\alpha$ (or $\vec{\alpha}$), and $b$ that minimize $H$.
Finally, the transition probability matrix $T$ is smoothed via interpolation with a uniform matrix $U$, such that $U_{ij} = 1\ /\left(l+u\right)$, and where $\epsilon$ is the interpolation parameter.
The smoothed transition matrix is defined as follows:
\begin{align}
	\tilde{T} = \epsilon U + \left(1 - \epsilon\right) T
\end{align}
The benefit of smoothing the transition probability matrix is best explained with an example. Let $\alpha=100$, $b = -100$: these parameters map every cosine similarity to the negative tail of the logistic function, where values approach 0 at an exponential rate. Further consider a word $x_1$ and its nearest neighbors $x_2$ and $x_3$, such that $cos\theta(x_1, x_2) = 0.8$ and $cos\theta(x_1,x_3) = 0.7$. Then, $w_{12} = 2.06\mathrm{e}{-9}$ and $w_{13} = 9.36\mathrm{e}{-14}$. Virtually all probability mass is on $w_{12}$ and labels only propagate to $x_1$ from $x_2$. The probability matrix needs to be smoothed to avoid this problem.

\subsubsection*{Label Propagation in batches}
Label propagation is a computationally efficient algorithm. Since iteration is avoided by directly computing the unique algebraic solution, the most computational resources are employed for the calculation of the probabilistic transition matrix $T$ and for the optimization of the parameters $\alpha$ (or $\vec{\alpha}$), $b$ and $\epsilon$.
Moreover, the size of $T$ can represent a memory issue. Consider that the Hashtag Corpus is comprised of $V=32,930$ word types. If we use the basic version of Label Propagation, i.e. that with one weight $\alpha \in \mathbb{R}$ shared by all dimensions, we only need to store the cosine similarity between each word pair so that $T \in \mathbb{R}^{V\times V}$. In this case, the transition matrix has a size of approximately 2GB for half-precision floating point numbers.  On the other hand, if we employ the hyperparameters $\vec{\alpha} \in \mathbb{R}^d$, $d$ elementwise products must be stored for each word pair as every product has to be scaled by a dimension-specific $\alpha$ (Equation \ref{eq:weight_sigmas}) at each epoch of the optimization.
The resulting matrix $T \in \mathbb{R}^{V \times V \times d}$ requires approximately 600GB for $d=300$.

To overcome this memory problem, we introduce Label Propagation in batches. Instead of keeping the entire $T \in R^{V \times V \times d}$ in memory during optimization, a subset of the vocabulary with size $W < V$ is randomly selected and the corresponding submatrix of $S_T \in \mathbb{R}^{W \times W \times d}$ is computed. If enough random submatrices are used for optimization, the obtained parameters will approximate those resulting from optimizing on $T \in R^{V \times V \times d}$.
Furthermore, the use of random submatrices is motivated by the need of the parameters to learn to adapt to any random subset of the vocabulary.

Randomly selecting $W$ word types can produce a skewed distribution of labeled and unlabeled instances: it is possible that a large amount of the word types are labeled, or that all words are unlabeled. Both these possibilities contradict the assumption of Label Propagation that $U \ll L$. Therefore, we fix the distribution of labeled and unlabeled instances to be equal to the proportion that they have in the original transition probability matrix.

Label propagation in batches can be used also for the optimization of $\alpha, b \in \mathbb{R}$ to reduce computation time.

\section{Corpus and lexicon analysis} \label{sec:analysis}
The Hashtag Emotion Corpus consists of 21,051 texts annotated with Ekman's six basic emotions. Each text is assigned a single emotion label. The corpus includes 32,929 word types.
The NRC Emotion Lexicon contains 14,182 words. However, only 3,462 lexicon words have at least one of Ekman's six emotion labels---the others are either annotated as \textit{positive}, \textit{negative}, \textit{anticipation}, \textit{trust}, or they are neutral, i.e. no label is set to 1 (Section \ref{relwork:resources}). Each lexicon word is tagged with an average of 0.44 Ekman's emotions. Table \ref{table:lab_per_lex} reports the labels-per-lemma statistics.

\begin{figure}
\centering
\begin{tikzpicture}
    \begin{axis}[
        ybar,
        symbolic x coords={anger, disgust, fear, joy, sadness, surprise},
        scale only axis,
        width=0.7\textwidth,
        height=0.4\textwidth,
        legend cell align={left},
        enlargelimits=0.06,
        legend style={at={(1,0.99)}, anchor=north east, draw=none}
    ]
    \addplot[ybar,fill=lightgray] coordinates {
        (anger, 1555) (disgust, 761) (fear, 2816) (joy, 8240) (sadness, 3830) (surprise, 3849)
        };
    \addplot[ybar, fill=teal] coordinates {
        (anger,1247) (disgust,1058) (fear,1476) (joy,689) (sadness,1191) (surprise,534)
        };
\legend{Hashtag Corpus,NRC Lexicon}
\end{axis}
\end{tikzpicture}
\caption{The class distributions of the Hashtag Corpus and the NRC Lexicon.}
\label{figure:class_dist}
\vspace{1ex}
\end{figure}
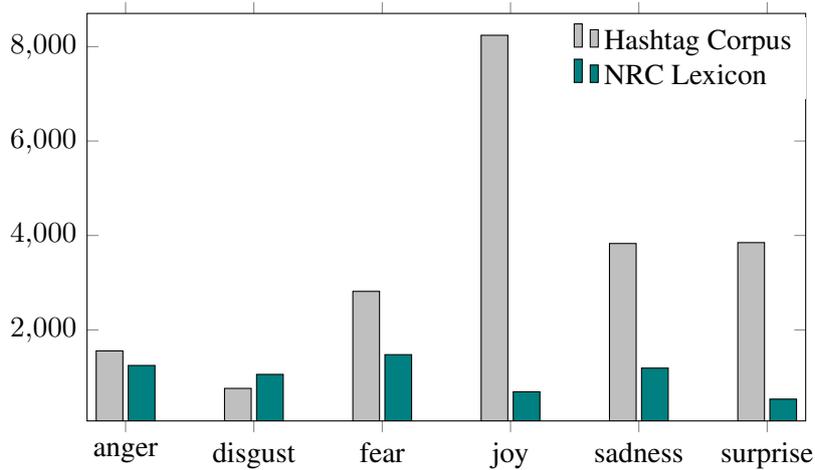

Furthermore, the class distributions are not uniform (Figure \ref{figure:class_dist}). In the lexicon, positive emotions (surprise and joy) are under-represented with respect to negative emotions (Table \ref{table:lexicon_class_dist}). In the corpus, texts annotated as \emph{joy} form a disproportionately large percentage, while \emph{anger} and \emph{disgust} are the minority classes (Table \ref{table:corpus_class_dist}).

\begin{table}[h!]
    \begin{minipage}[t]{.45\linewidth}
        \centering
        \begin{tabu} to 1\textwidth{ X[c] X[c] }
            \textbf{Emotion label} & \textbf{\# lemmas} \\
            \hline
            anger & 1247 \\
            disgust & 1058 \\
            fear & 1476 \\
            joy & \ \ 689 \\
            sadness & 1191 \\
            surprise & \ \ 534 \\
        \end{tabu}
        \caption{The class distribution of the NRC Emotion Lexicon.}
        \label{table:lexicon_class_dist}
    \end{minipage}
    \hspace{.08\linewidth}
    \begin{minipage}[t]{.45\linewidth}
        \centering
        \begin{tabu} to 1\textwidth{ X[c] X[c] }
        \textbf{Emotion label} & \textbf{\# texts}\\
        \hline
            anger & 1555 \\
            disgust & \ \ 761 \\
            fear & 2816 \\
            joy & 8240 \\
            sadness & 3830 \\
            surprise & 3849 \\
    \end{tabu}
    \caption{The class distribution of the Hashtag Corpus.}
    \label{table:corpus_class_dist}
    \end{minipage}
\vspace{1ex}
\end{table}

\begin{table}[h!]
    \begin{minipage}[t]{.45\linewidth}
        \centering
        \begin{tabu} to 1\textwidth{ X[c] X[c] }
            \textbf{\# labels} & \textbf{\# lemmas} \\
            \hline
            0 & 10720 \\
            1 & \ \ 1813 \\
            2 & \ \ \ \ 906 \\
            3 & \ \ \ \ 447 \\
            4 & \ \ \ \ 253 \\
            5 & \ \ \ \ \ \ 41 \\
            6 & \ \ \ \ \ \ \ \ 2 \\
            $>$ 0 & \ \ 3462 \\
        \end{tabu}
        \caption{The number of labels for a word in the NRC Emotion Lexicon.}
        \label{table:lab_per_lex}
    \end{minipage}
    \hspace{.07\linewidth}
    \begin{minipage}[t]{.47\linewidth}
        \centering
        \begin{tabu} to 0.97\textwidth{ X[c] X[c] }
            \textbf{\# lemmas} & \textbf{\# texts} \\
            \hline
            0 & 7513 \\
            1 & 7207 \\
            2 & 4187 \\
            3 & 1442 \\
            4 & \ \ 498 \\
            5 & \ \ 162 \\
            6 & \ \ \ \ 30 \\
            $>$ 6 & \ \ \ \ 12 \\
        \end{tabu}
        \caption{The number of emotion words in a text of the Hashtag Corpus.}
        \label{table:lex_per_txt}
    \end{minipage}
\end{table}

\begin{table}[h!]
    \centering
     \begin{tabular}{c c l}
        \textbf{Frequency} & \ \ \ \ \ \textbf{lemma} & \ \ \ \ \ \textbf{Labels}\\
        \hline
        1218 & \ \ \ \ \ love & \ \ \ \ \ \emph{joy}\\
        621 & \ \ \ \ \ good & \ \ \ \ \ \emph{joy, surprise}\\
        418 & \ \ \ \ \ afraid & \ \ \ \ \ \emph{fear}\\
        411 & \ \ \ \ \ happy & \ \ \ \ \ \emph{joy}\\
        389 & \ \ \ \ \ friend & \ \ \ \ \ \emph{joy}\\
        388 & \ \ \ \ \ god & \ \ \ \ \ \emph{fear, joy}\\
        367 & \ \ \ \ \ hate & \ \ \ \ \ \emph{anger, disgust, fear, sadness}\\
        342 & \ \ \ \ \ fear & \ \ \ \ \ \emph{anger, fear}\\
        311 & \ \ \ \ \ feeling & \ \ \ \ \ \emph{anger, disgust, fear, joy, sadness, surprise}\\
        287 & \ \ \ \ \ joy & \ \ \ \ \ \emph{joy}\\
    \end{tabular}
    \caption{The 10 most frequent NRC lemmas and their emotion labels.}
    \label{table:10most}
\end{table}

A text from the Hashtag Corpus contains on average 1.09 emotion words that also occur in the NRC lexicon and approximately one third of the tweets does not contain any. Table \ref{table:lex_per_txt} presents the distribution of emotion words among texts.
An emotion word has an average frequency of 6.64 and only 1,545 lemmas occur at least once in the Hashtag Corpus. Finally, Table \ref{table:10most} illustrates the most frequent emotion words along with their emotion labels according to the NRC Emotion Lexicon. The statistics provided in this paragraph were obtained by lemmatizing the Hashtag Corpus (the values resulting from the non-lemmatized corpus do not significantly vary).

The presented statistics clearly show that the coverage of an unexpanded lexicon is to small for emotion classification.

\section{Experiments} \label{sec:exp}
\subsection{Emotion-specific embeddings} \label{sec:exp-specific}
The first step of the proposed lexicon expansion method consists in learning emotion-specific word embeddings. These are distributed word representations that are able to encode affectual orientation and strength.
To learn an emotion-specific vector space we employ a recurrent neural network classifier. The classifier labels tweets from the Hashtag Corpus with Ekman's six basic emotions (Section \ref{relwork:resources}) and uses word vectors as trainable features. As the model learns to classify, we expect word embeddings to encode affectual orientation.

The proposed deep model is based on a bidirectional LSTM followed by a softmax or a sigmoid output layer. The emotion classifier is trained using Keras \citep{chollet2015keras} and the hyperparameters that we use are summarized in Appendix \ref{app:emo2vec}.
We experiment with different combinations of three regularization techniques: $\ell_2$  regularization, batch normalization, and dropout.
The choice between multinomial and multi-label classification determines the type of the output layer. A softmax output layer is used to obtain a probability distribution over the six emotion classes (multinomial classification). In contrast, a sigmoid layer is used to produce one probability value for each emotion class (multi-label classification).

For the initialization of word embeddings we rely on the vector space learned from the ENCOW corpus \citep{schafer2012cow,schafer2015cow} using the CBOW model. The ENCOW dataset contains approximately 425 million sentences and more than 9.5 billion tokens. The chosen vector dimensionality is $300$, as suggested by \citet{mikolov2013w2v}. We experiment with different thresholds of word frequency, excluding either words whose raw frequency in the corpus is lower than $100$ or those occurring less than $150$ times.
The context size is set to 5 as this training context was shown to give good performance in phrase analogy tasks \citep{mikolov2013distributed}. We expect this window size to be a desirable trade-off between computational complexity and the ability to capture semantic information.

The relevant output of the LSTM classifier are the optimized word embeddings that can be used to compute the word similarities necessary for lexicon expansion.

\subsection{Emotion lexicon expansion} \label{sec:exp-expansion}
To expand the NRC Emotion Lexicon we employ our novel variant of the Label Propagation algorithm. Although we have more than 300,000 word vectors at our disposal, label propagation is only applied to the approximately 30,000 vectors that correspond to the word types of the Hashtag Corpus. This decision is motivated by the need to limit the execution time of the propagation algorithm as well as the consideration that only the mentioned subset of word embeddings is optimized for emotion-related tasks. The word vectors are 300-dimensional.

As introduced in Section \ref{meth:expansion}, the label propagation problem can be solved with either one or multiple trainable hyperparameters.
We experiment with the scalar hyperparameters $\alpha,b \in \mathbb{R}$ and with $\alpha \in \mathbb{R}^{300}, b \in \mathbb{R}$. The label propagation hyperparameters are optimized using Tensorflow \citep{tensorflow2015} and their values are reported in Appendix \ref{app:labelprop}.

Furthermore, a batch-based variant of label propagation is introduced to overcome issues of excessive time and memory needs of hyperparameter optimization. Essentially, we try to approximate the results of standard label propagation optimization---where the word similarity graph is comprised of all available word types---with multiple batch optimizations---where only a subset of the word types is used to construct the similarity graph. We expect the parameters to learn to robustly adapt to any random subset of the vocabulary and, as a consequence, to discard irrelevant features.

\subsection{Emotion classification} \label{sec:exp-classification}
To test our hypothesis that a combination of corpus- and lexicon-based approaches improves classification, we use the expanded emotion lexicon to inform the classifier and to augment the word embeddings.

Our emotion classifier is the same as the one proposed for learning task-specific word embeddings. An embedding layer maps words to their vector representations. The embeddings are fed to a bidirectional LSTM, followed by a softmax or a sigmoid output layer, for multinomial and multi-label classification respectively. Again we experiment with $\ell_2$ regularization, batch normalization, and dropout.
The classifier is trained using Keras and its hyperparameters are reported in Appendix \ref{app:classification}.

For the initialization of word embeddings we leverage the vector space previously learned by our recurrent neural network model (Section \ref{sec:exp-specific}). This 300-dimensional vector space includes approximately 30,000 word embeddings. This is the corpus-based portion of the information we provide to the classifier.

To feed the classifier with lexicon-based information, we append the label probability distribution vector of a word occurring in the expanded emotion lexicon to the corresponding pretrained word vector.
As the lexicon is expanded to all the word types in the Hashtag Corpus, each embedding receives an emotion-specific initialization. Notice that the label distributions of the original lexicon words are left unvaried, due to the properties of Label Propagation.

\section{Evaluation and results} \label{sec:eval}
\subsection{Emotion-specific embeddings} \label{sec:eval-specific}
Evaluating the quality of the task-specific embeddings obtained via optimization of our emotion classifier is not a straightforward task. One could compute sample similarities between words to see if the embeddings capture our intuition about which words should be close to one another in the specialized vector space. Nevertheless, with the exception of very few indisputable cases, it is unfair to expect the embeddings to adhere to our judgments on the affective orientation of words if only because such judgments are inherently subjective. Consider, as an example, that it might feel natural to postulate a low similarity for \textit{happy} and \textit{sad} or a high similarity for \textit{scared} and \textit{terrified}. By doing so, however, we would implicitly reduce the dimensionality of the affective space to \textit{joy} / \textit{sadness}, or \textit{fear} respectively---in terms of Ekman's six basic emotions. Further consider an informal expression such as \textit{wtf}, which repeatedly occurs in the Hashtag Corpus. It is unclear what position it should occupy in a multidimensional affective space.

As we embed the optimization of specialized embeddings in a classification task, we decide to use the performance of the classifier as an extrinsic evaluation metric.
A second criterion is the performance of lexicon expansion, that we evaluate using 10-fold cross-validation.
The results of these evaluations are presented in the next two subsections.

\subsection{Emotion lexicon expansion} \label{sec:eval-expansion}
To perform an intrinsic evaluation of our lexicon expansion method, we choose 10-fold cross-validation. The intersection between the word types of the Hashtag Corpus and the NRC Emotion Lexicon is partitioned into 10 equal sized subsamples. Cross-validation is repeated 10 times and each of the subsamples is used once as validation data.

The quality of the expanded lexicon is assessed by computing the average Kullback-Leibler divergence between the emotion label probability distributions obtained normalizing the NRC Emotion Lexicon and the the distributions resulting from label propagation.

Three baseline lexicon expansion methods are introduced: (i) assigning a uniform class distribution to all words, (ii) assigning to all words a distribution where all the probability mass is given to the majority class based on the Hashtag Corpus, (iii) assigning to all words the prior class distribution of the Hashtag Corpus.

Table \ref{table:kl} shows the average Kullback--Leibler divergence for 10-fold cross-validation of the described lexicon expansion techniques.
\begin{table}[h!]
    \begin{center}
        \begin{tabular} { l  c }
            \textbf{Lexicon expansion} & \textbf{KL divergence} \\
            \hline
            Uniform distribution & 1.34 \\
            Majority class (Hashtag Corpus) & 21.32 \\
            Prior class distribution (Hashtag Corpus) & 1.53 \\
            Label propagation  ($\alpha \in \mathbb{R}$)  & 1.31 \\
            Batch label propagation  ($\alpha \in \mathbb{R}$)  & 1.31 
        \end{tabular}
    \end{center}
    \caption{Average Kullback--Leibler divergence for 10-fold cross-validation on the NRC Emotion Lexicon. The lowest divergence is obtained using label propagation.}
    \label{table:kl}
\end{table}

Interestingly, the uniform class distribution yields an even lower divergence than the prior class distribution of the Hashtag Corpus. Although its low divergence, the uniform distribution clearly cannot be used in practice as it is, by definition, uninformative.

Label propagation with a scalar parameter $\alpha$ is the method that best minimizes the average Kullback--Leibler divergence, indicating that the quality of the expanded lexicon is satisfactory. Remarkably, batch label propagation performs similarly to standard propagation although it uses batches of size 5,000 (compared to a total of more than 30,000 word types). This result shows that batch approximation can work for graph propagation, at least when the number of trainable parameters is limited.

In contrast, using batch approximation to optimize the parameter vector $\vec{\alpha}$ appears to require either a large batch size, or a large number of training batches, both of which demand a long runtime. Optimization performed with 500 batches of size 3,000 (each batch is trained on for 5 epochs) yields an average Kullback--Leibler divergence of 14.37. Increasing the batch size to 4,000 and the number of batches to 1,500 (each batch is trained on for 3 epochs), the average Kullback--Leibler divergence drops to 13.23.
We can therefore conclude that a parameter vector $\vec{\alpha}$ is ineffective if is optimized using batch gradient descent. It remains unclear if $\vec{\alpha}$ is generally inadequate for this task or if its optimization requires the entire dataset. Therefore, we will only report the classification results based on label propagation with a scalar parameter $\alpha$.

\subsection{Emotion classification} \label{sec:eval-classification}
To evaluate the proposed emotion classifier, we introduce multiple baselines.
The lower bound is represented by a random classifier. Then we implemented a trivial count-based classifier that assigns emotion labels based on the NRC Emotion Lexicon. In particular, this trivial classifier counts the lexicon words occurring in a paragraph for each emotion class---$fr\left( e_k \right)$---, and produces a label probability distribution over $m$ classes:
\[
    P\left( e_i \right) = \frac{fr\left( e_i \right)}{\sum_{k=1}^{m} fr\left( e_k \right)}
\]

A more competitive baseline is represented by a version of the LSTM classifier introduced at the beginning of this section that only exploits our emotion-specific word embeddings.
As a final alternative, the pretrained emotion-specific embeddings are concatenated with the probability distributions indicated by the unexpanded NRC Emotion Lexicon. Words that do not occur in the lexicon have their vectors concatenated with a vector of probabilities randomly sampled from a uniform distribution. Uniform initialization outperforms the overall probability distribution of emotions in the lexicon.

The precision, accuracy, and $F_1$ score of all the presented classifiers are reported in Tables \ref{table:classify-hashtag} and \ref{table:classify-semeval}. All metrics are computed using micro-averaging to allow a comparison with previous work and to reduce the effect of label imbalance. Using macro-averaging would assign more weight to the majority classes, for which classifiers tend to perform better due to the larger amount of training instances.
\begin{table}[h!]

    \begin{center}
        \begin{tabular} { l c c c }
            \textbf{Classifier} & \textbf{P} & \textbf{R} & \textbf{F1}\\
            \hline
            Random & 16.9 & 16.9 & 16.9 \\
            Count-based (NRC Emotion Lexicon) & 15.7 & 15.7 & 15.7 \\
            One-vs-all SVM \citep{mohammad2015hashtag} & 55.1 & 45.6 & 49.9 \\
            Multinomial LSTM & 55.0 & 55.0 & 55.0 \\
            Multinomial LSTM + NRC Emotion Lexicon & 55.2 & 55.2 & 55.2 \\
            Multinomial LSTM + expanded lexicon ($\alpha \in \mathbb{R}$) & \textbf{56.2} & \textbf{56.2} & \textbf{56.2} \\
            \hline
            Students & 40.9 & 40.4 & 40.6
        \end{tabular}
    \end{center}
    \caption{Results of classification on the Hashtag Emotion Corpus. Although the bidirectional LSTM classifier represents a strong baseline, using the expanded lexicon boosts classification accuracy.}
    \label{table:classify-hashtag}
\end{table}

For the Hashtag Emotion Corpus, the bidirectional LSTM classifier introduced as a baseline outperforms one-vs-all SVM with binary features, setting a relatively high lower bound to our task. Including the label distributions of the NRC Emotion Lexicon as features slightly increases the classifier accuracy, already indicating that corpus-based and lexicon-based information is complementary. The limited increment in accuracy can be explained by the fact that a text from the Hashtag Corpus includes on average 1.09 NRC emotion words and that approximately one third of the tweets does not contain any NRC lemmas.
The LSTM classifier shows a remarkable increase in accuracy when the expanded lexicon is provided. Although we can assume that quality of the expanded lexicon is lower than the quality of the hand-annotated NRC Emotion Lexicon, the wider coverage of the former seems to successfully help the classifier.

Regardless of whether the LSTM classifier uses no lexicon, the NRC lexicon, or the expanded one, the multinomial variant consistently obtains a higher accuracy. The classification report of our best classifier is presented in Table \ref{table:classify-best}.
\begin{table}[h!]
    \small
    \begin{center}
        \begin{tabular} { l c c c }
            \textbf{Classifier} & \textbf{P} & \textbf{R} & \textbf{F1}\\
            \hline
            One-vs-all SVM \citep{mohammad2015hashtag} \\
            \tabitem 1. ngrams in headlines dataset and Hashtag Corpus + domain adaptation & 46.0 & 35.5 & 40.1 \\
            \tabitem 2. ngrams in headlines dataset + NRC Emotion Lexicon & \textbf{46.7} & 38.6 & 42.2 \\
            Multi-label LSTM & 38.8 & 50.3 & 43.8 \\
            Multi-label LSTM + NRC Emotion Lexicon & 39.2 & \textbf{50.9} & 44.3 \\
            Multi-label LSTM + expanded lexicon & 43.1 & 48.9 & \textbf{45.9} \\
        \end{tabular}
    \end{center}
    \caption{Results of classification on the SemEval headlines dataset. The most accurate classifier is the bidirectional LSTM informed with the expanded lexicon.}
     \label{table:classify-semeval}
\end{table}

The fact that label propagation is performed on ENCOW embeddings with backpropagation from the supervised learning task possibly suggests that the expanded lexicon is tailored to the training dataset (Hashtag Corpus). However, the performance of the LSTM classifier is boosted by the expanded lexicon even on a dataset from a different domain (SemEval headlines).
This result seems to imply that the Hashtag Corpus is a sufficiently large resource, which grants coverage over a wide variety of word types and encodes context-independent information.

\subsubsection*{Humans as classifiers}
In Section \ref{relwork:emoclass}, we have reported the inter-annotator agreement studies conducted by \citet{strapparava2007semeval} for the SemEval headlines corpus. These have shown that trained annotators agree with a Pearson correlation of 53.67 using the simple average over classes, and $r = 43$ using the frequency-based average.

Instead of reproducing another inter-annotator agreement study, we test the accuracy of an untrained person with respect to an emotion-annotated dataset, the Hashtag Corpus. Our survey includes 33 participants. These are undergraduate and graduate students asked to read 25 tweets and to classify them into one of Ekman's emotion classes. Each participant is given a different set of tweets, for a total of 825 classified instances. Not all the students are native English speakers. The results of our survey are shown in Table \ref{table:classify-students}.

In the attempt to establish an upper bound, we find evidence that an untrained annotator---in other words, an average person---is considerably less accurate than all our LSTM classifiers.
This outcome seems to confirm our hypothesis that the accuracy of the state-of-the-start emotion classifiers, although it appears very low compared to that of e.g. valence classifiers, is close to some yet unknown upper bound.
Assigning an emotion to a short paragraph is a hard task for both a human and a statistical classifier as it requires more contextual information than it is available in the paragraph itself.

\begin{table}[h!]
	\begin{minipage}[t]{.45\linewidth}
		\centering
		\begin{tabu} to 1\textwidth{ X[l] X[c] X[c] X[c] }
			\textbf{Class} & \textbf{P} & \textbf{R} & \textbf{F1} \\
			\hline
			anger & 25 & 50 & 33 \\
			disgust & 18 & 70 & 29 \\
			fear & 48 & 22 & 30 \\
			joy & 52 & 46 & 49 \\
			sadness & 50 & 52 & 51 \\
			surprise & 40 & 23 & 29 \\
			\hline
			average & \ \ \ 40.9 & \ \ \ 40.4 & \ \ \ 40.6
		\end{tabu}
		\caption{Results of classification on the Hashtag Emotion Corpus, students.}
		\label{table:classify-students}
	\end{minipage}
	\hspace{.08\linewidth}
	\begin{minipage}[t]{.45\linewidth}
		\centering
		\begin{tabu} to 1\textwidth{ X[l] X[c] X[c] X[c]  }
			\textbf{Class} & \textbf{P} & \textbf{R} & \textbf{F1}\\
			\hline
			anger & 38 & 27 & 32 \\
			disgust & 40 & 18 & 25 \\
			fear & 58 & 52 & 55 \\
			joy & 66 & 76 & 71  \\
			sadness & 40 & 44 & 42  \\
			surprise & 53 & 46 & 49 \\
			\hline
			average & \ \ \ 56.2 & \ \ \ 56.2 & \ \ \ 56.2
		\end{tabu}
		\caption{Results of classification on the Hashtag Emotion Corpus, best classifier: bidirectional LSTM with expanded lexicon.}
		\label{table:classify-best}
	\end{minipage}
	\vspace{1ex}
\end{table}

\begin{figure}
\centering
\begin{tikzpicture}
    \begin{axis}[
        ybar,
        symbolic x coords={anger, disgust, fear, joy, sadness, surprise},
        scale only axis,
        width=0.7\textwidth,
        height=0.4\textwidth,
        enlargelimits=0.06,
        legend cell align={left},
        legend style={at={(1,0.99)}, anchor=north east, draw=none}
    ]
    \addplot[ybar,fill=lightgray] coordinates {
        (anger, 7.39) (disgust, 3.62) (fear, 13.38) (joy, 39.14) (sadness, 18.19) (surprise, 18.28)
        };
    \addplot[ybar, fill=teal] coordinates {
        (anger, 17.01) (disgust, 12.42) (fear, 6.03) (joy, 32.57) (sadness, 18.82) (surprise, 11.46)
        };
\legend{Hashtag Corpus,Survey}
\end{axis}
\end{tikzpicture}
\caption{The class distributions in the Hashtag Corpus and in the answers to the survey (true positives + false positives). The values on the y-axis are percentages.}
\label{figure:class_dist_survey}
\vspace{1ex}
\end{figure}
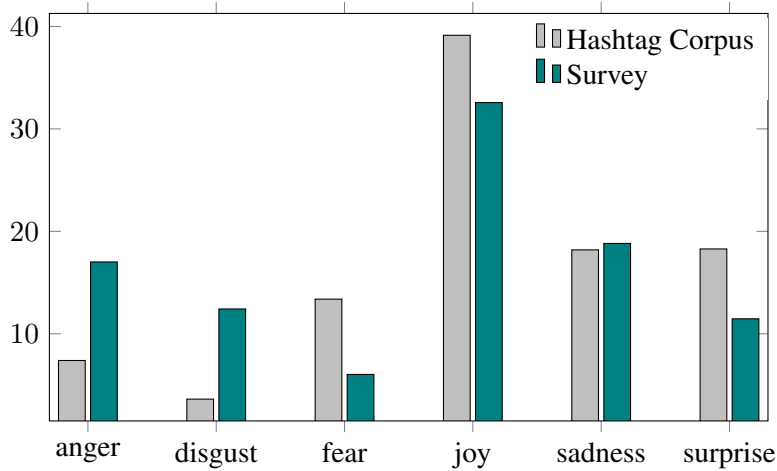

A possible explanation for the surprisingly low accuracy of the average person is to be found in the quality of the Hashtag Corpus. \citet{mohammad2015hashtag} clarify that there are essentially three types of tweets: those where the affectual orientation is straightforward even without the final hashtag, those where only the hashtag makes the affectual orientation explicit, and those where text and hashtag seem to conflict. In the second and in the third case, humans tend to answer according to a uniform prior while the model's prior belief corresponds to the class distribution of the dataset. The accuracy of the model seems to benefit from such an informed prior distribution.

A final consideration is that, in the absence of clear emotional content, both humans and the model are expected to comply with the gold standard label selected by annotators whose agreement score is remarkably low. It is plausible that, for such complex cases, there exist simply no universally correct emotion label.

\section{Conclusion} \label{sec:concl}
In this thesis, we have argued and shown that combining the corpus-based and the lexicon-based approaches can improve the accuracy of emotion classifiers.

In particular, we have shown that label propagation can expand an emotion lexicon in a meaningful way and that graph propagation can rely on task-specific word embeddings. We have introduced two variations of the Label Propagation algorithm: (i) a novel weight computation that allows the use of cosine similarity as a distance metric, (ii) and a batch-based training algorithm which reduces the time and memory needed to propagate labels throughout very large graphs.

Moreover, we have presented a method to learn emotion-specific word embeddings from a corpus of emotion-annotated short paragraphs: such specialized word vectors are the result of the optimization of an an emotion classifier.
We have proposed a bidirectional LSTM classifier that proved to be accurate even without additional information from a lexicon. Finally, we have shown that feeding the class probability distributions learned via label propagation to our classifier improves its performance.

The software related to this paper is open-source and available at \url{https://github.com/Procope/emo2vec}.

In future, we want to test if good optimization of multiple label propagation parameters can be performed using even larger number of batches and batch size. We also plan to employ GloVe embeddings as an initialization for trainable specialized vectors as they were shown to capture semantic similarity better than Word2Vec embeddings \citep{pennington2014glove}.
Furthermore, we want to introduce lexical-contrast information into our task-specialization routine using wordnets, and to---instead of optimizing word embeddings singularly---learn a rotation that can transform the entire vector space into a task-specific one.

Emotion-specific embeddings have proven to be a reliable source for the construction of similarity graphs but there may be other interesting options: we will experiment with co-occurrence counts and with wordnets in order to discover alternative meaningful word representations.

	\newpage
	\bibliography{thesis}
	\bibliographystyle{acl_natbib}

	\newpage
	\appendix
	\section{Hyperparameters}\medskip

    \subsection{Emotion-specific embeddings}\label{app:emo2vec}
    The emotion classifier was trained using Keras \citep{chollet2015keras}. Here, we provide an
    overview of the hyperparameters that we used.
    \begin{itemize}
    	\item \textbf{Solver}: Adagrad, with a learning rate decay of $1\mathrm{e}{-4}$.
    	\item \textbf{Learning rate}: The initial learning rate is set to $0.005$.
    	\item  \textbf{Epochs}: The model was trained for 30 epochs.
    	\item  \textbf{LSTM layer}: The forward and backward layers were trained with 128 output dimensions. Increasing the number of output dimensions did not provide an improvement.
    	\item \textbf{Regularization}: 10\% dropout and 20\% recurrent dropout \cite{srivastava2014dropout}. A stronger dropout did not provide better performance. Additionally, we applied $\ell_2$ regularization (Keras default).
    \end{itemize}\medskip

    \subsection{Label propagation} \label{app:labelprop}
    The parameters of our variant of Label Propagation were optimized using Tensorflow \cite{tensorflow2015}.
    \begin{itemize}
    	\item \textbf{Standard label propagation}
    	\begin{itemize}
    		\item epochs: $100$
    		\item $\alpha = 0.007$
    		\item $b = 2.41$
    	\end{itemize}
    	\item \textbf{Batch-based label propagation}
    	\begin{itemize}
    		\item batch size: $5000$
    		\item number of batches: $1000$
    		\item epochs per batch: $3$
    		\item $\alpha = -0.001$
    		\item $b = 0.9$
    	\end{itemize}
    \end{itemize} \medskip

    \subsection{Emotion classification} \label{app:classification}
    This is an overview the hyperparameters that were used for the best emotion classifier.
	\begin{itemize}
		\item \textbf{Solver}: Adagrad, with a learning rate decay of $1\mathrm{e}{-3}$.
		\item \textbf{Learning rate}: The initial learning rate is set to $0.01$.
		\item  \textbf{Epochs}: The model was trained for 20 epochs.
		\item  \textbf{LSTM layer}: The forward and backward layers were trained with 128 output dimensions. Increasing the number of output dimensions did not provide an improvement.
		\item \textbf{Regularization}: 10\% dropout and 20\% recurrent dropout \cite{srivastava2014dropout}. A stronger dropout did not provide better performance. Additionally, we applied $\ell_2$ regularization (Keras default).
	\end{itemize}

\end{document}